\documentclass[runningheads]{llncs}

\PassOptionsToPackage{hidelinks,pdfborder={0 0 0}}{hyperref}
\usepackage{hyperref}
\hypersetup{
  colorlinks=false,
  pdfborder={0 0 0},
  pdfborderstyle={/S/U/W 0} 
}

\usepackage[T1]{fontenc}

\usepackage{graphicx}
\usepackage[export]{adjustbox}

\usepackage{subcaption} 

\usepackage{amsmath,amssymb,amsfonts}

\usepackage{algorithm}         
\usepackage{algorithmicx}
\usepackage{algpseudocode}

\usepackage{url}
\usepackage{wrapfig}
\usepackage{multirow}
\usepackage{booktabs}
\usepackage{array}
\usepackage{cellspace}
\usepackage{threeparttable}
\usepackage{multicol}
\usepackage{tikz}
\usepackage{pgfplots}
\usepackage{soul}
\usepackage{xcolor}
\usepackage{orcidlink}
\usepackage{bbding}    
\usepackage{colortbl}
\usepackage{filecontents}
\usepackage{cite}

\usepackage{marvosym}  

\usepackage[utf8]{inputenc}

\definecolor{CBL}{RGB}{107,209,187}
\definecolor{SMN}{RGB}{217,84,117}
\definecolor{SBN}{RGB}{243,154,143}
\definecolor{LIN}{RGB}{248,229,194}
\definecolor{VAN}{RGB}{186,208,119}
\definecolor{DMN}{RGB}{125,166,198}
\definecolor{FPN}{RGB}{91,114,136}
\definecolor{DAN}{RGB}{224,199,227}
\definecolor{VN}{RGB}{157,128,99}
\definecolor{color1}{rgb}{0.98,0.81,0.69}
\definecolor{color2}{rgb}{0.55,0.71,0.0}
\definecolor{color3}{rgb}{1.0,0.6,0.4}
\definecolor{color4}{rgb}{0.29,0.59,0.82}
\definecolor{color5}{HTML}{b4b4b4}
\definecolor{color6}{HTML}{5f97d2}
\definecolor{color7}{HTML}{c76d72}

\DeclareUnicodeCharacter{2019}{'}
\DeclareUnicodeCharacter{2032}{\ensuremath{'}}   
\DeclareUnicodeCharacter{00B4}{'}                

\begin{document}
\title{Structure Matters: Brain Graph Augmentation via Learnable Edge Masking for Data-efficient Psychiatric Diagnosis}
\titlerunning{Structure Matters: Brain Graph Augmentation via Learnable Edge Masking}
%
\author{Mujie Liu\inst{1}\orcidlink{0009-0002-0879-7168}\and
Chenze Wang\inst{2}\orcidlink{0009-0005-6558-6764}\and
Liping Chen\inst{3}\orcidlink{0009-0001-0937-4290}\and
Nguyen Linh Dan Le\inst{3}\orcidlink{0009-0008-5888-0932}\and 
Niharika Tewari\inst{3}\orcidlink{0000-0002-7690-9814}\and
Ting Dang \inst{4}\orcidlink{0000-0002-7932-4571} \and
Jiangang Ma \inst{1}\orcidlink{0000-0002-8449-7610} \and
Feng Xia\inst{3} \textsuperscript{(\Letter)}  \orcidlink{0000-0002-8324-1859 }}
\authorrunning{M. Liu et al.}
%
\institute{Institute of Innovation, Science and Sustainability, Federation University Australia, Ballarat 3353, Australia \\ \and
School of Information and Electronic Engineering, Zhejiang Gongshang University, Hangzhou 315104, China \\ \and
School of Computing Technologies, RMIT University, Melbourne 3000, Australia\\ \and School of Computing and Information Systems, The University of Melbourne, Melbourne 3052, Australia  \\  \email{mujie.liu@ieee.org}, \email{23020090112@pop.zjgsu.edu.cn}, 
\email{lp.chen@ieee.org}, \email{s4135772@student.rmit.edu.au}, 
\email{S4044597@student.rmit.edu.au}, \email{ting.dang@unimelb.edu.au}, \email{j.ma@federation.edu.au}, \email{f.xia@ieee.org}}

%

\maketitle              

\begin{abstract}
The limited availability of labeled brain network data makes it challenging to achieve accurate and interpretable psychiatric diagnoses. While self-supervised learning (SSL) offers a promising solution, existing methods often rely on augmentation strategies that can disrupt crucial structural semantics in brain graphs.
To address this, we propose SAM-BG, a two-stage framework for learning brain graph representations with structural semantic preservation. In the pre-training stage, an edge masker is trained on a small labeled subset to capture key structural semantics. In the SSL stage, the extracted structural priors guide a structure-aware augmentation process, enabling the model to learn more semantically meaningful and robust representations. 
Experiments on two real-world psychiatric datasets demonstrate that SAM-BG outperforms state-of-the-art methods, particularly in small-labeled data settings, and uncovers clinically relevant connectivity patterns that enhance interpretability. Our code is available at {\url{https://github.com/mjliu99/SAM-BG}.}

\end{abstract}
\section{Introduction}

Functional magnetic resonance imaging (fMRI) is a widely adopted neuroimaging technique in neuroscience research. By capturing blood-oxygen-level-dependent (BOLD) signals, fMRI enables functional connectivity measurement, allowing researchers to monitor brain activity and assess neurological development~\cite{yang2020current}. Increasing evidence shows that disruptions in brain network connectivity are closely linked to the onset and progression of various neurological and psychiatric disorders~\cite{li2024complexity}. These findings underscore the importance of developing advanced computational methods tailored to fMRI data to better support clinical diagnostics and mental health care through physiologically grounded brain network analysis~\cite{amemiya2024resting}.

Graph learning~\cite{gllongsurvey2025arxiv} has emerged as a powerful tool in this context, representing fMRI data as brain graphs where nodes correspond to brain regions and edges represent functional connections. Leveraging graph neural networks (GNNs), these methods effectively capture complex connectivity patterns and have shown strong potential in tasks such as disease classification and biomarker discovery~\cite{10388338, peng2024learning}. However, their success often depends on large-scale labeled datasets, which are scarce in psychiatric research due to privacy concerns and the high cost of expert annotations. For example, labeling patients with autism spectrum disorder (ASD) typically requires long-term behavioral observation and developmental assessments conducted by trained clinicians~\cite{hyman2020executive}. 

To alleviate this challenge, self-supervised learning (SSL) has emerged as a promising paradigm for learning meaningful representations from unlabeled data~\cite{xie2022self, liu2022graph}. SSL-based methods typically rely on generic graph augmentation strategies such as random edge perturbation, node masking, or subgraph sampling to generate multiple views for contrastive or predictive learning~\cite{zhang2023gcl, peng2024adaptive, luo2024interpretable, wang2025self}. While effective in general graph domains, these augmentation methods are task-agnostic and domain-independent, often neglecting domain-specific structural priors. In brain network analysis, such priors are particularly crucial, as they encode biologically meaningful structural semantics, often manifested as connectivity patterns (substructures), which are closely associated with psychiatric disorders~\cite{gallo2023functional}. Thus, indiscriminate perturbations of brain connections in current augmentation strategies can distort essential structural semantics in the resulting graph views, leading to unrealistic and less explainable representations. This distortion ultimately undermines the effectiveness of SSL-based methods in downstream clinical and diagnostic applications. 
    
To address this issue, we propose SAM-BG, a framework designed for low-supervision settings that uses structure-aware augmentation via edge masking to learn robust, biologically meaningful brain graph representations from large-scale, unlabeled fMRI data. It follows a two-stage learning strategy: (i) \emph{Structure Semantic Extraction}, which pre-trains an edge masker using a small set of labeled data to identify the most discriminative substructures, guided by the information bottleneck (IB) principle~\cite{yu2021recognizing}. (ii) \emph{SSL with Structure-semantic Preservation}, where the trained edge masker is applied to unlabeled brain graphs to retain key substructures while introducing controlled perturbations to the remaining graph. This ensures that augmented views preserve critical semantic information while adding sufficient variability. Through this design, SAM-BG improves both representation robustness and interpretability, offering insights into disorder-relevant connectivity patterns. Our main contributions are:

    \begin{itemize}
    \item We propose a novel two-stage learning framework, SAM-BG, which improves data efficiency for psychiatric disorder diagnosis by leveraging limited supervision to learn structurally meaningful representations from large-scale, unlabeled brain networks. 
    
    \item We introduce a learnable masker that enhances the learning of biologically meaningful structural semantics, which in turn improves data augmentation for more effective representation learning in psychological applications.

    \item Extensive experiments on two real-world fMRI datasets demonstrate that SAM-BG outperforms state-of-the-art methods in low-data scenarios and reveals disorder-specific brain connectivity patterns.


    

    \end{itemize}

\vspace{-15pt}

\section{Related Work}

\subsection{Brain Network Analysis on fMRI Data}
The fMRI data captures brain connectivity through BOLD signal fluctuations and is widely used in brain network analysis. Traditional methods~\cite{plitt2015functional, pan2018novel} typically flatten this data into feature vectors, ignoring the brain’s topological structure, an essential aspect for detecting connectivity disruptions in neurological disorders. Graph learning addresses this limitation by representing fMRI data as graphs. BrainGB~\cite{cui2022braingb} provides a unified benchmark framework for evaluating graph learning methods on brain networks. Other models, such as BrainGNN~\cite{li2021braingnn}, employ region-aware pooling to identify key regions of interest (ROIs), thereby enhancing disease-related interpretability. In contrast, IBGNN~\cite{cui2022interpretable} highlights important ROIs and connections through a learnable global mask. 
However, these methods heavily depend on labeled data, which is often limited in psychiatric research. To address this challenge, we propose SAM-BG that improves representation learning from unlabeled fMRI data by preserving structural-semantic information during SSL.

\subsection{Self-supervised Brain Network Representation Learning}

SSL has recently emerged as a powerful approach for learning representations without relying on labeled data, effectively addressing data scarcity~\cite{liu2022graph}. Through data augmentation strategies, SSL enables the extraction of discriminative features. 
In graph-based SSL, methods like GraphCL~\cite{you2020graph} generate positive pairs through edge perturbation, node dropping, and subgraph sampling. For brain networks, A-GCL~\cite{zhang2023gcl} and SF-GCL~\cite{peng2024stage} adopt Bernoulli masking for edge-level perturbations, while BraGCL~\cite{luo2024interpretable} improves interpretability by incorporating node and edge importance. GCDA~\cite{wang2025self} further introduces a diffusion-based pretext task to enhance semantic augmentation. 
However, many existing methods neglect structural semantics during augmentation, leading to biased representations. To overcome this, we propose a structure-aware graph augmentation strategy that generates biologically meaningful views for learning more robust and explainable representations.

\section{The Proposed Method}
\begin{figure*}[t!]
        \centering
        \includegraphics[width=\textwidth]{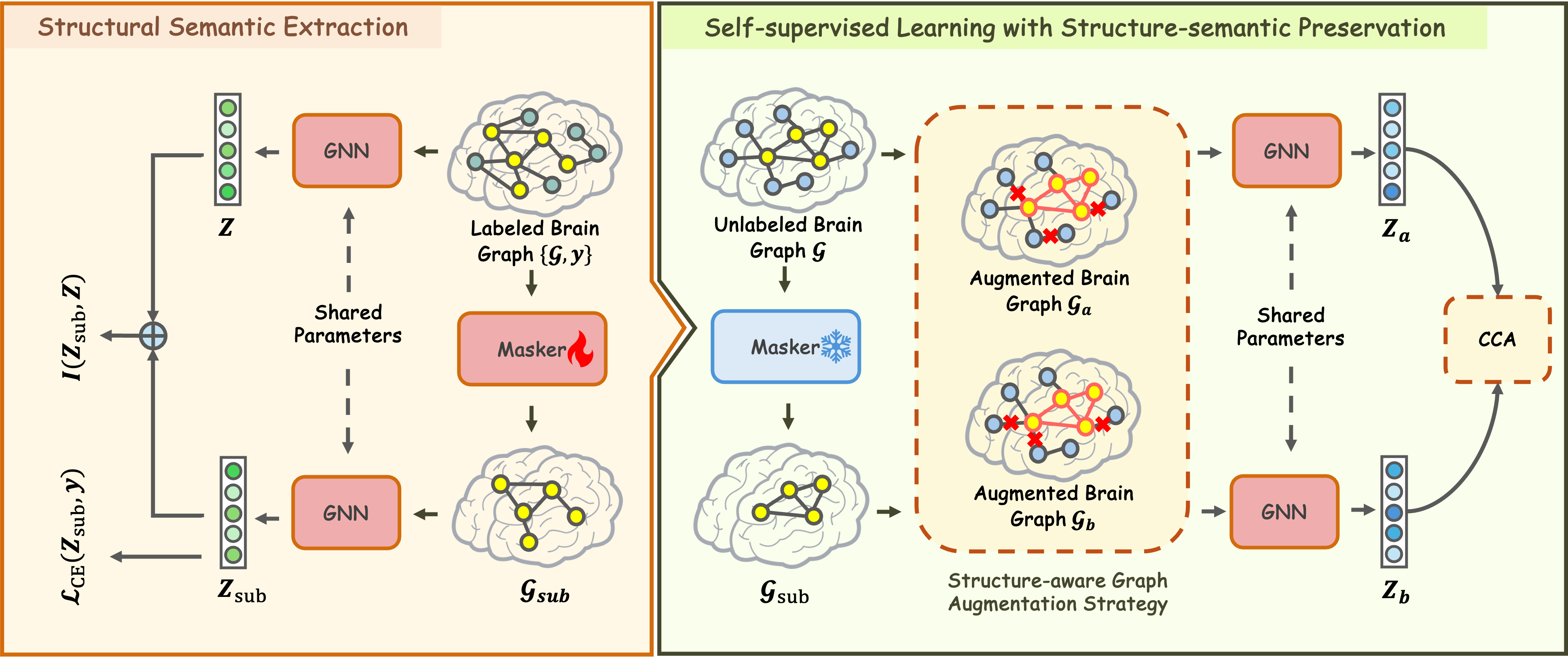} 
        \caption{Overview of the SAM-BG model consisting of structural semantic extraction and SSL with structure-semantic preservation phases.}
        \label{fig: framework}
    \end{figure*}

\vspace{-5pt}
    The overall framework of SAM-BG is shown in Figure~\ref{fig: framework}. It consists of two main phases:
    (i) Structural semantic extraction, where an edge masker is trained to identify substructures containing task-relevant structural semantics; 
    (ii) SSL with structure-sematic preservation, where augmentations are applied only to non-salient edges while retaining the extracted substructures, thus enhancing brain network representations by maintaining critical structural semantics. 

    \subsection{Problem Definition}

    We represent a brain network as a graph \( \mathcal{G} = (\mathcal{V}, \mathcal{E}, \mathbf{A}, \mathbf{X}) \), where \( \mathcal{V} = \{v_1, v_2, \dots, \\ v_N\} \) denotes the set of \( N \) ROIs, \( \mathbf{A} \in \mathbb{R}^{N \times N} \) is the adjacency matrix computed via Pearson correlation~\cite{schober2018correlation}, and \( \mathbf{X} \in \mathbb{R}^{N \times D} \) is the node feature matrix with each row \( \mathbf{x}_i \in \mathbb{R}^D \) representing the feature vector of ROI \( v_i \). An edge \( e_{i,j} \in \mathcal{E} \) exists if \( a_{i,j} \in \mathbf{A}\) indicates significant functional connectivity. 
    Given a small labeled dataset \( \{\mathcal{G}^i, y^i\}_{i=1}^{S} \) and a large unlabeled set \( \{\mathcal{G}^j\}_{j=1}^{U} \), where \( U \gg S \) and \( y \in \{0,1\} \), our goal is to learn biologically meaningful brain graph representations \( \mathbf{Z} \in \mathbb{R}^{N \times D} \), which are further used for downstream tasks.

    \subsection{Structural Semantic Extraction}
    
\begin{figure*}[t!]
        \centering
        \includegraphics[width=\textwidth]{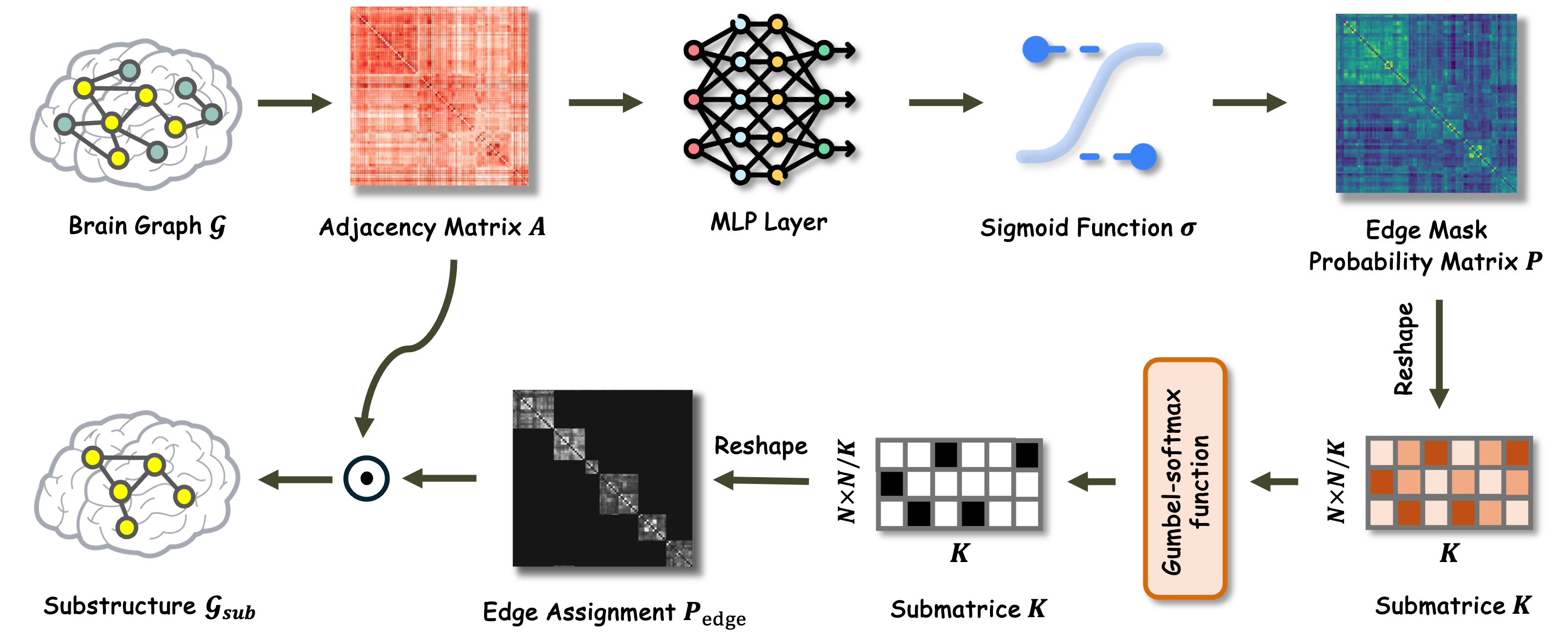} 
        \caption{Framework of the subgraph generation process. The input graph $\mathcal{G}$ is processed by the masker to sample a subset of edges and construct a subgraph $\mathcal{G}_{\text{sub}}$ that preserves key structural information.}
        \label{fig: masker}
    \end{figure*}

    Since brain connections (edges) are pivotal for diagnosing and interpreting the neural mechanisms of psychiatric disorders~\cite{li2020neuroimaging, gallo2023functional}, we introduce an edge masker to distill discriminative substructures as structural priors (Figure~\ref{fig: masker}).
    For a given labeled brain network \( \mathcal{G} \), the edge masker operates directly on \( \mathbf{A} \), encoding it via a two-layer multi-layer perceptron (MLP) network followed by a sigmoid activation function to generate edge selection probabilities \( \mathbf{P} \in \mathbb{R}^{N \times N} \). Each element $p_{i,j} \in [0,1]$ denotes the selection probability of edge $e_{i,j}$. To impose sparsity while avoiding over-pruning, we divide the probability matrix \( \mathbf{P} \) into \( K \) submatrices, each containing \( {N \times N}/{K} \) edges. Within each group, the Gumbel–Softmax reparameterization~\cite{maddison2017concrete} selects one edge among \( K \) consecutive candidates, yielding a differentiable, structurally coherent mask. The grouped masks are reshaped back to \( N \times N \) size and binarized to form the final edge selection mask $\mathbf{P}_{\text{edge}}$. The substructure of brain network \(\mathcal{G}_{\text{sub}} \) is obtained by element-wise masking of the original graph: \(\mathbf{A}_{\text{sub}} = \mathbf{A} \odot \mathbf{P}_{\text{edge}}\).

    Then, the masker is trained under the Information Bottleneck (IB) principle~\cite{yu2021recognizing}, as illustrated in Figure~\ref{fig: framework} (left), aiming to retain label-relevant semantics while discarding redundant topology. Specifically, the objective is to minimize the mutual information (MI) between $\mathcal{G}_{\text{sub}}$ and $\mathcal{G}$ to encourage compression, while maximizing the MI between $\mathcal{G}_{\text{sub}}$ and the label $y$ to ensure discriminability. The objective is formally expressed as:
    
    \begin{equation}
        \min_{\mathcal{G}_{\text{sub}}} \,  \beta I(\mathcal{G}_{\text{sub}}, \mathcal{G}) - I(\mathcal{G}_{\text{sub}}, y),
    \end{equation}    

    \noindent where $\beta$ is the hyperparameters, and $I(\cdot,\cdot)$ is the MI estimation.
    To compute the first term, we encode the embeddings of $\mathcal{G}_{\text{sub}}$ and $\mathcal{G}$ using a shared GNN encoder as $\mathbf{Z}_{\text{sub}}$ and $\mathbf{Z}$. We then use the matrix-based R{\'e}nyi's \(\alpha\)-order MI estimator \cite{yu2019multivariate} to obtain their MI as $I(\mathcal{G}_{\text{sub}}, \mathcal{G}) := I(\mathbf{Z}_{\text{sub}}, \mathbf{Z})$.
    For the second term, we follow previous work~\cite{yu2021recognizing} that minimizes the cross-entropy loss between the $\mathbf{Z}_{\text{sub}}$ and its ground truth label $y$ as $-I(\mathcal{G}_{\text{sub}}, y):=\mathcal{L}_{\text{CE}}(\mathbf{Z}_{\text{sub}}, y)$.
    Thus, the final objective becomes:
    \begin{equation}
    \label{loss_pre}
        \mathcal{L}_{\text{pre}} =  \beta I(\mathbf{Z}_{\text{sub}}, \mathbf{Z}) +  \mathcal{L}_{\text{CE}}(\mathbf{Z}_{\text{sub}}, y).
    \end{equation}


    This formulation encourages the edge masker to extract compact, label-relevant substructures $\mathcal{G}_{\text{sub}}$ that preserve essential structural semantics, performing as the structural prior knowledge. 

    \subsection{Self-supervised Learning with Structure-sematic Preservation}

    As shown in Figure~\ref{fig: framework} (right), we propose an SSL-based representation learning module, which incorporates structural priors to mitigate the information loss commonly introduced by random perturbations in existing methods. The framework comprises a structure-aware graph augmentation strategy with a tailored learning objective to learn semantically meaningful representations for data-efficient diagnosis of psychiatric disorders. 
    
    \paragraph{Structure-aware Graph Augmentation Strategy.}
    An effective augmentation should increase diversity without compromising semantic integrity. To this end, we integrate a trained edge masker to transfer structural priors from limited labeled data to large-scale unlabeled data, thereby generating biologically plausible augmented brain graphs that support meaningful representation learning. Specifically, the masker first extracts salient substructures \( \mathcal{G}_{\text{sub}} \) from unlabeled brain graphs. To introduce variability, we perturb the complementary non-salient part \( \Delta \mathcal{G} = \mathcal{G} \setminus \mathcal{G}_{\text{sub}} \) by applying uniform edge dropout to its edge set \( \mathcal{E}_{\Delta} \). For each edge \( e_{i,j} \in \mathcal{E}_{\Delta} \), an independent mask is sampled as \( m_{i,j} \sim \mathrm{Bernoulli}(1-\epsilon) \), where \( \epsilon \) is the drop rate; edges in the salient subgraph are always preserved, i.e., \( m_{i,j}=1 \) for \( e_{i,j} \in \mathcal{G}_{\text{sub}} \). For brain networks, symmetry is enforced by setting \( m_{i,j}=m_{j,i} \). The augmented adjacency is then updated element-wise as \( \mathbf{A}' = \mathbf{A} \odot \mathbf{M} \) (equivalently, \( a'_{i,j} = a_{i,j} m_{i,j} \)). Resampling the masks yields two complementary yet semantically aligned views,\(\mathcal{G}_a=(\mathcal{V}, \mathcal{E}_a, \mathbf{A}_a, \mathbf{X}) \quad \text{and} \quad \mathcal{G}_b=(\mathcal{V}, \mathcal{E}_b, \mathbf{A}_b, \mathbf{X})\), both preserving core semantic structure while providing sufficient diversity.

    \paragraph{Structure-aware Representation Learning Objective.}

    After generating structurally meaningful augmented brain graphs, we adopt an SSL framework to learn graph-level representations. To capture shared structural semantics across views, SAM-BG employs canonical correlation analysis (CCA)~\cite{andrew2013deep} as its core learning objective. CCA encourages the model to learn perturbation-invariant yet semantically aligned embeddings by maximizing the correlation between the augmented view representations. Specifically, the two augmented brain graphs, $\mathcal{G}_a$ and $\mathcal{G}_b$, are fed into a shared GNN encoder, which is initialized using the encoder pre-trained during the structure masker phase. This encoder generates view-specific embeddings $\mathbf{Z}_a$ and $\mathbf{Z}_b$, which are then optimized via a CCA-based loss to align the representations in the latent space. The learning objective is defined as:

    \begin{equation}
		\label{Loss_CCA}
		\mathcal{L}_{\text{CCA}} =  \left\| \mathbf{Z}_a- \mathbf{Z}_b \right\|_{\text{F}}^2 +   \lambda \left( \left\| \mathbf{Z}_a^{\top}\mathbf{Z}_a - \mathbf{I} \right\|_{\text{F}}^2 + \left\| \mathbf{Z}_b^{\top}\mathbf{Z}_b - \mathbf{I} \right\|_{\text{F}}^2 \right ),
	\end{equation}
    where the $\lambda$ is the trade-off coefficient, and $\mathbf{I}$ is an identity matrix. The first term of the objective aligns the embeddings of the two augmented graphs in a shared latent space, promoting perturbation-invariant representations. The second term enforces decorrelation across embedding dimensions, preserving view-specific features and preventing representation collapse. 
    By jointly optimizing these two terms, the trained GNN encoder can effectively learn meaningful and discriminative representations $\mathbf{Z}$ from any given brain network $\mathcal{G}$. These representations can then be applied to various downstream tasks, such as psychiatric disorder classification or biomarker discovery. 
    The overall training procedure of SAM-BG is summarized in Algorithm~\ref{algorithm}.    



    \begin{algorithm}[h]
    \caption{Training Procedure of the SAM-BG Model}
    \label{algorithm}
    \begin{algorithmic}[1]
        \State \textbf{Input:} 
            Labeled brain network dataset $\{\mathcal{G}^{i}, y^{i}\}_{i=1}^S$, 
            Unlabeled brain network dataset $\{\mathcal{G}^{j}\}_{j=1}^U$, 
            Training epochs $T$
        \State \textbf{Output:} 
            Optimized model parameters $\hat{\theta}_{\text{GNN}}, \hat{\theta}_{\text{Masker}}$ 
        \State \textbf{Initialize:} 
            Model parameters $\theta_{\text{GNN}}, \theta_{\text{Masker}}$
        
        \State \textit{Phase I: Structural Semantic Extraction Phase (supervised)}
        \For{epoch $= 1$ to $T$}
            \For{each labeled sample $i = 1$ to $S$}
                \State Compute edge mask probability $\mathbf{P}^{i} = f_{\theta_{\text{Masker}}}(\mathcal{G}^{i})$
                \State Binarize $\mathbf{P}^{i}$ to obtain substructure $\mathcal{G}_{\text{sub}}^{i}$
                \State Obtain representations $\mathbf{Z}^{i} = f_{\theta_{\text{GNN}}}(\mathcal{G}^{i})$ and $\mathbf{Z}_{\text{sub}}^{i} = f_{\theta_{\text{GNN}}}(\mathcal{G}_{\text{sub}}^{i})$
                \State Compute loss $\mathcal{L}_{\text{pre}}$ (Equation~\ref{loss_pre})
                \State Update $\theta_{\text{GNN}}$, $\theta_{\text{Masker}}$ via gradient descent
            \EndFor
        \EndFor
        \State \textbf{Return:} Trained $\hat{\theta}_{\text{Masker}}$ and warm-start ${\theta}_{\text{GNN}}$ 
        
        \State \textit{Phase II: Self-supervised Representation Learning Phase}
        \For{epoch $= 1$ to $T$}
            \For{each unlabeled sample $j = 1$ to $U$}
                \State Extract substructure $\mathcal{G}_{\text{sub}}^{j}$ using trained masker $f_{\hat{\theta}_{\text{Masker}}}(\mathcal{G}^{j})$
                \State Construct augmented graphs $\mathcal{G}_a^{j}$ and $\mathcal{G}_b^{j}$ by perturbing $\Delta \mathcal{G}^{j} = \mathcal{G}^{j} \backslash \mathcal{G}_{\text{sub}}^{j}$
                \State Obtain representations $\mathbf{Z}_a^{j} = f_{\theta_{\text{GNN}}}(\mathcal{G}_a^{j})$ and $\mathbf{Z}_b^{j} = f_{\theta_{\text{GNN}}}(\mathcal{G}_b^{j})$
                \State Compute loss $\mathcal{L}_{\text{CCA}}$ (Equation~\ref{Loss_CCA})
                \State Update $\theta_{\text{GNN}}$ via gradient descent
            \EndFor
        \EndFor
        
        \State \textbf{Return:} $\hat{\theta}_{\text{GNN}}, \hat{\theta}_{\text{Masker}}$ 
    \end{algorithmic}
    \end{algorithm}

\vspace{-20pt}
\section{Experiment}
    \subsection{Experiment Setup}
    \paragraph{Datasets.} 
    We conducted experiments using two real-world medical datasets: Autism Brain Imaging Data Exchange (ABIDE)~\footnote[1]{\url{https://fcon_1000.projects.nitrc.org/indi/abide/}} and Attention Deficit Hyperactivity Disorder (ADHD-200)~\footnote[2]{\url{https://fcon_1000.projects.nitrc.org/indi/adhd200/}} datasets. 
    Both contain resting-state fMRI data commonly used in psychiatric research. For ABIDE, we selected a subset of 743 participants, comprising 364 ASD patients (ages 7–24) and 379 HC individuals (ages 6–22), with balanced age and gender distribution. For ADHD-200, we utilized a sample of 459 participants from this dataset, consisting of 229 typically developing (TD) individuals and 230 children and adolescents diagnosed with ADHD (ages 7–21).

    \paragraph{Data Preprocessing} 
    We preprocess resting\mbox{-}state fMRI into graph\mbox{-}ready brain networks. The pipeline includes slice\mbox{-}timing correction, head\mbox{-}motion correction, spatial normalization, and Gaussian smoothing, implemented with the Graph Theoretical Network Analysis (GRETNA) toolbox\footnote[3]{\url{https://www.nitrc.org/projects/gretna/}} and SPM12\footnote[4]{\url{https://www.fil.ion.ucl.ac.uk/spm/software/spm12/}}. The brain is then parcellated using the Automated Anatomical Labeling (AAL) atlas, which defines \(116\) ROIs as nodes. For each ROI, we extract the BOLD time series and compute pairwise Pearson correlations~\cite{schober2018correlation} between ROI signals to weight their functional connectivity, resulting in a weighted functional brain network for each subject.
 
    \paragraph{Baselines.}
    We compare SAM-BG with several representative baselines, including supervised and SSL methods. For supervised learning, we adopt BrainGNN\cite{li2021braingnn} and BrainGB\cite{cui2022braingb}, where BrainGNN employs an ROI-aware pooling mechanism, and BrainGB provides a standardized benchmark for brain graph analysis. For SSL, we include four competitive methods: GraphCL\cite{you2020graph}, GCA\cite{zhu2021graph}, MA-GCL\cite{gong2023ma}, and CCA-SSG\cite{zhang2021canonical}. GraphCL explores several graph augmentation strategies; GCA improves this by adapting to salient graph structure; MA-GCL generates views based on learned embeddings; and CCA-SSG uses CCA to capture perturbation-invariant features without contrastive pairs. 
    For a fair comparison, we adopt the official settings for BrainGNN and BrainGB. All SSL baselines use their original implementations with edge perturbation and 20\% labeled data for fine-tuning, simulating low-supervision scenarios.


    \paragraph{Implementation Details.} Our model is implemented using PyTorch and trained on an NVIDIA Tesla P100 GPU. During pre-training, we leverage 20\% of the labeled data to guide masker training. For downstream tasks, the dataset is split into 80\% for fine-tuning, 10\% for validation, and 10\% for testing, with the test set strictly excluded from any supervised training during pre-training. The hyperparameters are configured as follows: the MI trade-off coefficient $\beta$ is set to 0.01 for ABIDE and 0.001 for ADHD, while the weight of the decorrelation loss term $\lambda$ is set to $1\text{e}^{-4}$ for ABIDE and 0.001 for ADHD. To balance sparsity and structural completeness, the number of submatrices is set to $K = N/2$. The GNN encoder adopts graph isomorphism networks with bilinear second-order pooling to enhance node interaction and reduce parameters. 
    We conduct 5-fold cross-validation and repeat experiments five times with different random seeds. Results are reported as mean and standard deviation of accuracy (ACC), the area under the ROC curve (AUC), recall, and F1-score.

\subsection{Experimental Results and Discussion}

   \begin{table*}[t!]
        \centering
        \small  
        \caption{Performance (\%) on the ABIDE and ADHD datasets. SL represents supervised learning, and SSL represents self-supervised learning. The best results are bold, and the second results are underlined. }
        \label{tab1}
        \setlength{\extrarowheight}{2pt} 
        \resizebox{\linewidth}{!}{  
        \begin{tabular}{c|c|cccc|cccc}
        \toprule
            &\multirow{2}{*}{\textbf{Methods}}  &  \multicolumn{4}{c|}{\textbf{ABIDE}} & \multicolumn{4}{c}{\textbf{ADHD}} \\
            \cmidrule{3-10}
            && \textbf{ACC} & \textbf{AUC}  & \textbf{Recall}  & \textbf{F1-score} &  \textbf{ACC} & \textbf{AUC}  & \textbf{Recall}  & \textbf{F1-score} \\
            \midrule
            \multirow{2}{*}{\textbf{SL}} & BrainGNN  & $54.9\pm 8.7$ & $57.1 \pm 8.3$ & $55.0 \pm 6.3$ & $53.4 \pm 6.0$ & $55.8 \pm 1.2$ & $58.0 \pm 4.9$ & $46.3 \pm 4.2$ & $55.9 \pm 4.2$ \\
            &BrainGB  & ${55.6\pm 4.3}$ & $56.3 \pm 3.6$ & ${55.4 \pm 5.2}$ & ${53.6 \pm 10.2}$ & $56.2 \pm 7.1$ & $59.6 \pm 5.2$ & $56.2 \pm 4.6$ & ${61.6 \pm 6.3}$ \\
            \midrule
            \multirow{6}{*}{\textbf{SSL}}
             &MA-GCL  &$49.1\pm 7.1$ & $44.8 \pm 9.2$ & $52.5 \pm 7.6$ & $42.5 \pm 11.8$  &$50.1\pm 5.6$ & $48.5 \pm 6.2$ & $46.5 \pm 3.7$ & $45.2 \pm 6.4$ \\
            &GraphCL  &$51.3\pm 4.6$ & $50.4 \pm 5.7$ & $52.4 \pm 7.5$ & $50.3 \pm 6.4$  &$51.5\pm 6.8$ & $47.6 \pm 5.9$ & $47.8 \pm 7.5$ & $46.5 \pm 5.6$ \\
            &GCA  &$52.8\pm 3.5$ & $50.6 \pm 2.4$ & $53.2 \pm 4.4$ & $52.1 \pm 2.3 $ &$50.8 \pm 2.4$ &$52.4 \pm 5.1$ &$59.4 \pm 3.6$ &$53.6 \pm 4.4$\\
            & CCA-SSG   &  $55.0\pm 6.1$ & ${57.2 \pm 6.0}$ & $55.2 \pm 7.7$ & $53.3 \pm 6.7$ &${56.6 \pm 5.8}$ &${60.3 \pm 2.6}$ &${60.3 \pm 4.4}$ &${58.0 \pm 8.6}$\\
            &{BraGCL} & $  {53.5\pm 4.3}$ & $ {55.1 \pm 5.2}$ & $  {53.4 \pm 4.4}$ & $  {51.8 \pm 5.0}$ & $  {63.8 \pm 5.6}$ & $  {67.3 \pm 3.7}$ & $  {63.8 \pm 5.6}$ & $  {63.7 \pm 5.5}$ \\
            & {A-GCL}   & $  \underline{59.0\pm 1.6}$ & $  {\underline{63.3 \pm 1.8}}$ & ${\mathbf{71.6 \pm 12.9}}$ & $  \underline{63.0 \pm 4.1}$ & $  {\underline{64.4 \pm 2.5}}$ & $  {\mathbf{69.0 \pm 3.2}}$ & $  {\underline{68.5 \pm 17.2}}$ & $  {\underline{64.7 \pm 3.9}}$ \\
        \cmidrule{2-10}
            \cmidrule{1-10}
            &\textbf{SAM-BG} (ours)  & $\mathbf{63.1\pm 3.4}$ & $\mathbf{66.7\pm 4.3}$ & $\underline{65.9\pm 4.2}$ & $\mathbf{66.7\pm 5.2}$ &$\mathbf{67.7 \pm 3.7}$ &$\underline{66.5 \pm 4.6}$ &$\mathbf{68.8 \pm 3.8}$ &$\mathbf{68.4 \pm 5.4}$\\
            \bottomrule
        \end{tabular}
        }\label{Performance comparison}
    \end{table*}

  \begin{table*}[t!]
    \centering
    \tiny  
    \caption{Performance (\%) of the ablation study for substructure evaluation on the ABIDE and ADHD datasets.}
    \label{table2}
    \setlength{\extrarowheight}{2pt} 
    \resizebox{\linewidth}{!}{  
    \begin{tabular}{c|cccc|cccc}
        \hline
        \multirow{2}{*}{\textbf{Methods}}  &  \multicolumn{4}{c|}{\textbf{ABIDE}} & \multicolumn{4}{c}{\textbf{ADHD}} \\
        \cline{2-9}
        & \textbf{ACC} & \textbf{AUC}  & \textbf{Recall}  & \textbf{F1-score} &  \textbf{ACC} & \textbf{AUC}  & \textbf{Recall}  & \textbf{F1-score} \\
        \hline
        SAM-BG (sub)   & ${50.1\pm 4.3}$ & ${51.4 \pm 5.2}$ & $55.6 \pm 3.9$ & $51.8 \pm 5.5$ & $51.3 \pm 3.4$ & $52.5 \pm 6.2$ & $53.3 \pm 5.6$ & ${52.9 \pm 5.1}$ \\
        SAM-BG (rest) &$46.4\pm 4.5$ & $44.7 \pm 4.3$ & $48.1 \pm 5.1$ & $44.7 \pm 4.8$  &$48.3\pm 4.8$ & $53.7 \pm 5.1$ & $49.5 \pm 3.2$ & $50.9 \pm 6.2$ \\
        SAM-BG (only) & ${58.3\pm 5.6}$ & ${61.3\pm 4.6}$ & ${57.8\pm 5.7}$ & ${58.5\pm 7.3}$ &${61.3 \pm 4.3}$ &${60.2 \pm 4.2}$ &${61.5 \pm 7.4}$ &${59.5 \pm 6.5}$\\
        \hline
    \end{tabular}
    } 
    \end{table*}

    \subsubsection{Comparison Results.} 
    Table~\ref{tab1} presents the complete comparative results between our method and the baseline methods on the ABIDE and ADHD datasets. Overall, supervised learning methods tend to outperform most SSL methods in terms of accuracy, highlighting the importance of label information. Among them, BrainGB attains the highest F1-score, indicating strong robustness under full supervision. Despite this, CCA-SSG achieves competitive results across several metrics, benefiting from its non-contrastive objective, which facilitates better preservation of structural-semantic information. In particular, our proposed method, SAM-BG, consistently outperforms both state-of-the-art supervised and SSL methods across the two datasets. On ABIDE, it improves ACC by 4.1\%, AUC by 3.4\%, and F1-score by 3.7\% compared to the best baseline, although its recall is slightly lower than A-GCL. On ADHD, it achieves 3.3\% higher ACC, 0.3\% higher recall, and 3.7\% higher F1-score, with a marginally lower AUC than A-GCL. Notably, SAM-BG demonstrates more substantial improvements on the ABIDE dataset, suggesting that disease-related connectivity patterns may be more discriminative in autism brain networks.
    We attribute the superiority of SAM-BG to its two-stage training strategy, which uses limited labeled samples to guide structurally meaningful representation learning on large-scale, unlabeled brain networks.

    \subsubsection{Ablation Study.}
 
    In this section, we evaluate the effectiveness of the structure-aware strategy and the semantic quality of extracted substructures, demonstrating our model's overall robustness and reliability.

   \paragraph{The Effectiveness of Extracted Substructure From Masker.}
    To evaluate the role of the identified substructures, we compare three variants: (i) perturbing only the selected substructures (SAM-BG(sub)), (ii) removing them and perturbing the rest of the graph (SAM-BG(rest)), and (iii) using only the substructures for classification (SAM-BG(only)). As shown in Table~\ref{table2}, SAM-BG(rest) performs worst when the substructures are removed, while SAM-BG(only) is closest to the original model. This observation indicates that the identified substructures encode the brain network's most discriminative and label-relevant semantic features. Disrupting these substructures shifts the model’s attention toward semantically irrelevant structural information, degrading learned representations and impairing downstream performance. These findings empirically validate the effectiveness of our structural masker in isolating non-essential components and preserving core topological semantics. Moreover, they highlight the potential of our method to guide the model toward learning biologically meaningful and task-relevant representations.

    \begin{figure}[t!]

        \centering
        \begin{subfigure}[b]{0.48\textwidth} 
            \centering
            \includegraphics[width=\textwidth]{  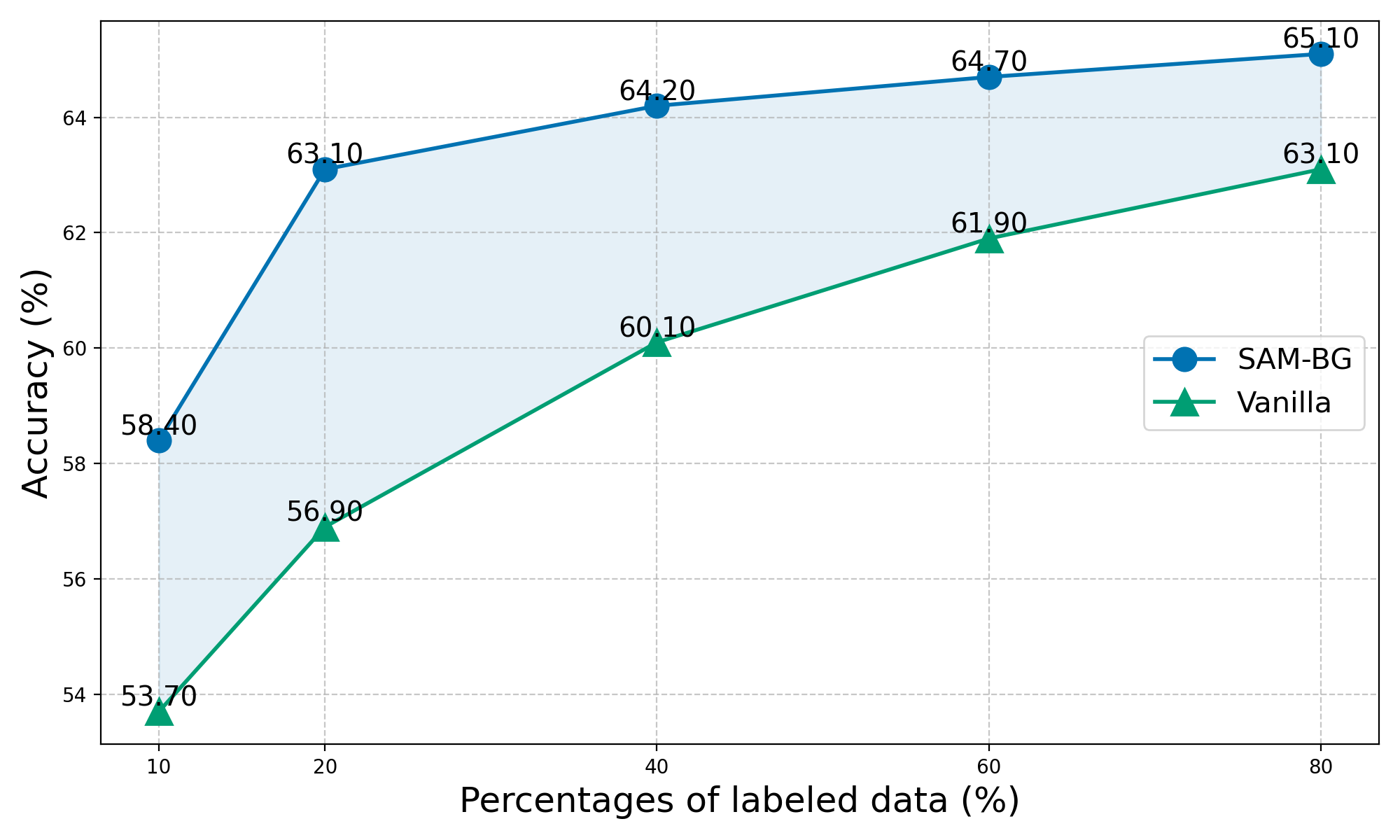}
            \caption{ABIDE}
        \end{subfigure}
        \begin{subfigure}[b]{0.48\textwidth} 
            \centering
            \includegraphics[width=\textwidth]{  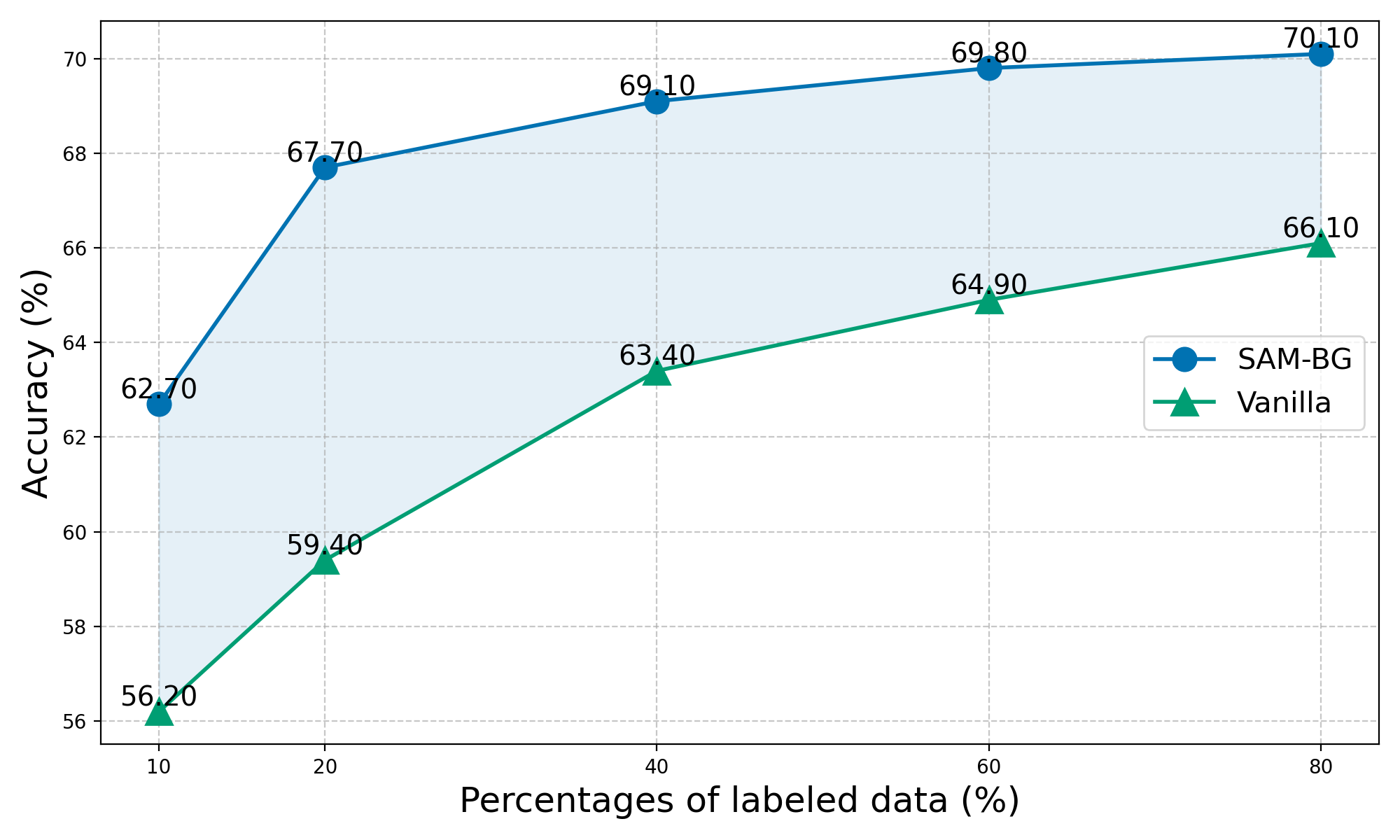}
            \caption{ADHD}
        \end{subfigure}

        \caption{Comparison between SAM-BG and Vanilla model with the increasing number of labeled data.}
        \label{fig: fine-tunning}
    \end{figure}

    \paragraph{The Effectiveness of Structure-aware Graph Augmentation Strategy in Data-efficient Settings.}
    To evaluate the impact of the learnable edge masking strategy in our augmentation process, we compared our model to a vanilla baseline that uses random structural perturbations under varying proportions of labeled data. As shown in Figure \ref{fig: fine-tunning}, our approach consistently outperforms the baseline on both datasets, with accuracy improving as the amount of labeled data increases. Notably, with 20\% labeled data, our model achieves the highest gains over the baseline, with improvements of 6.2\% and 8.3\%, respectively, and delivers performance comparable to the vanilla model trained with 80\% labeled data. Furthermore, the results highlight the superior performance of our method in low-label scenarios, where it outperforms most existing SSL approaches even with only 10\% labeled data. These findings validate the effectiveness of incorporating structural-semantic priors into the representation learning process. They also highlight the model's ability to deliver reliable performance even under limited supervision.

	\begin{figure}[t!]
        \centering
        
        \begin{subfigure}[b]{0.48\textwidth} 
            \centering
            \includegraphics[width=\textwidth]{  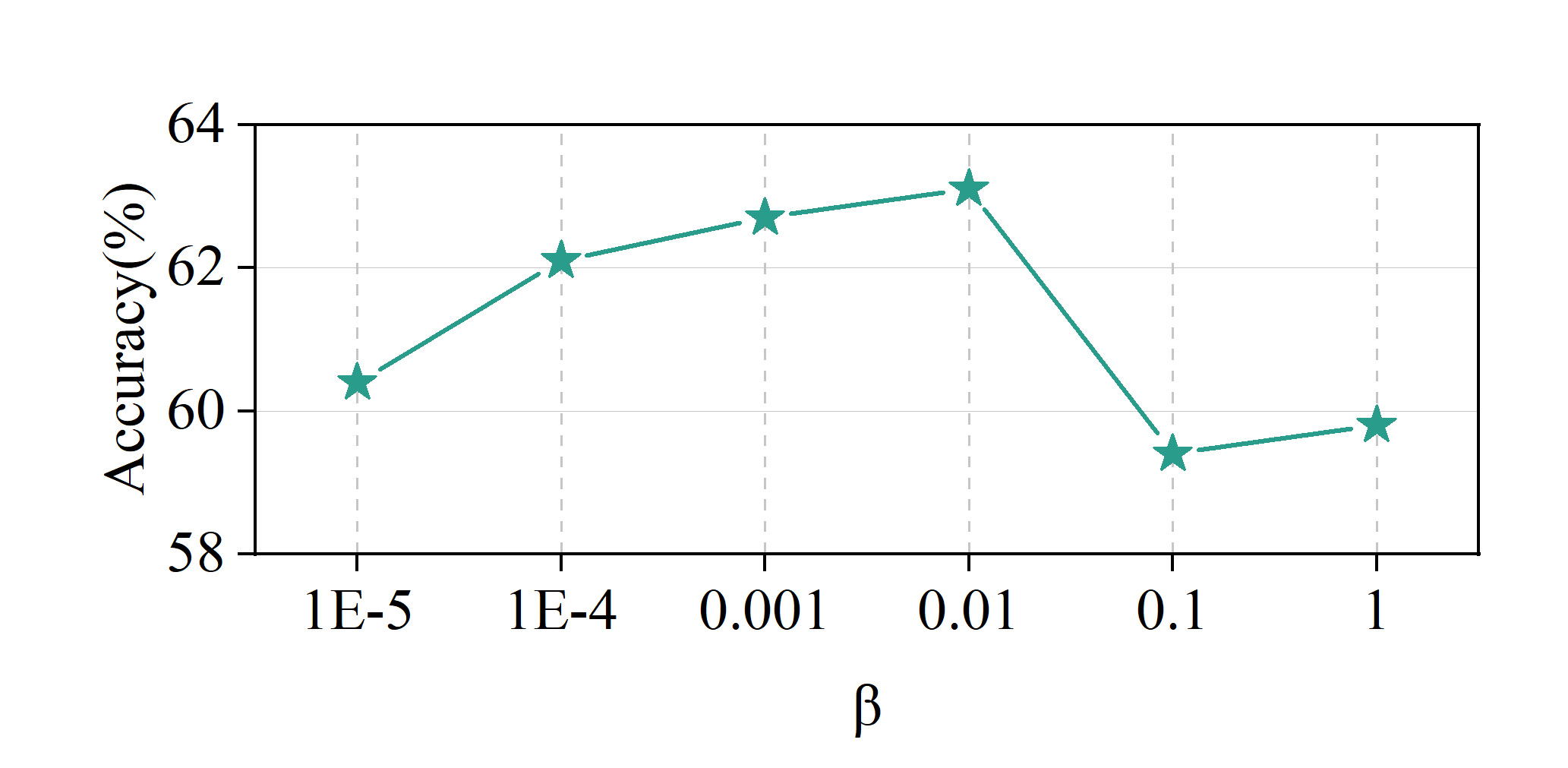}
            \caption{ABIDE}
        \end{subfigure}
        \begin{subfigure}[b]{0.48\textwidth} 
            \centering
            \includegraphics[width=\textwidth]{  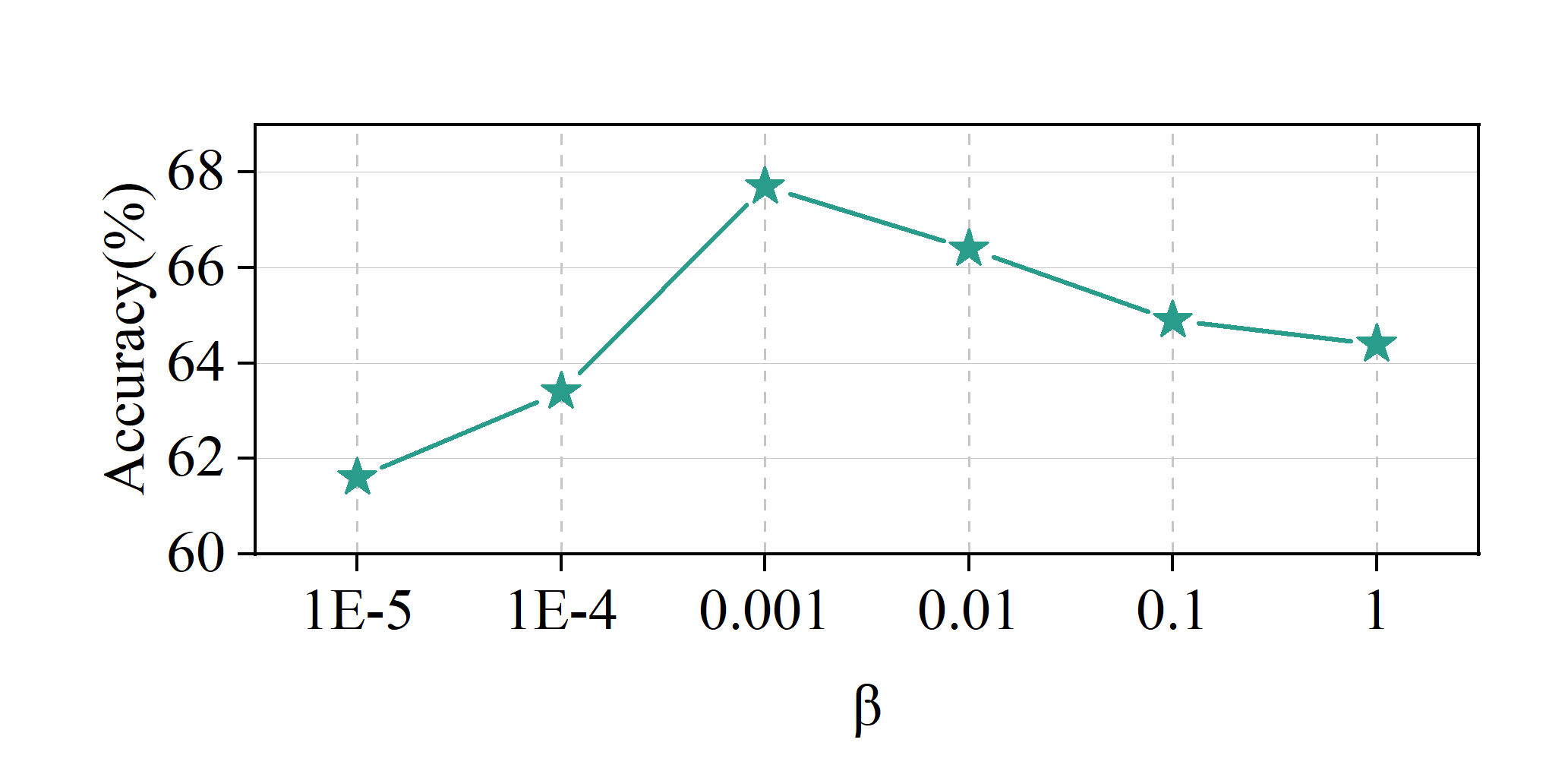}
            \caption{ADHD}
        \end{subfigure}
        \caption{Accuracy (\%) of SAM-BG with different values of $\beta$ in Equation \ref{loss_pre}.}
        \label{fig: beta}
    \end{figure}

    \begin{figure}[t!]
        \centering
        
        \begin{subfigure}[b]{0.48\textwidth} 
            \centering
            \includegraphics[width=\textwidth]{  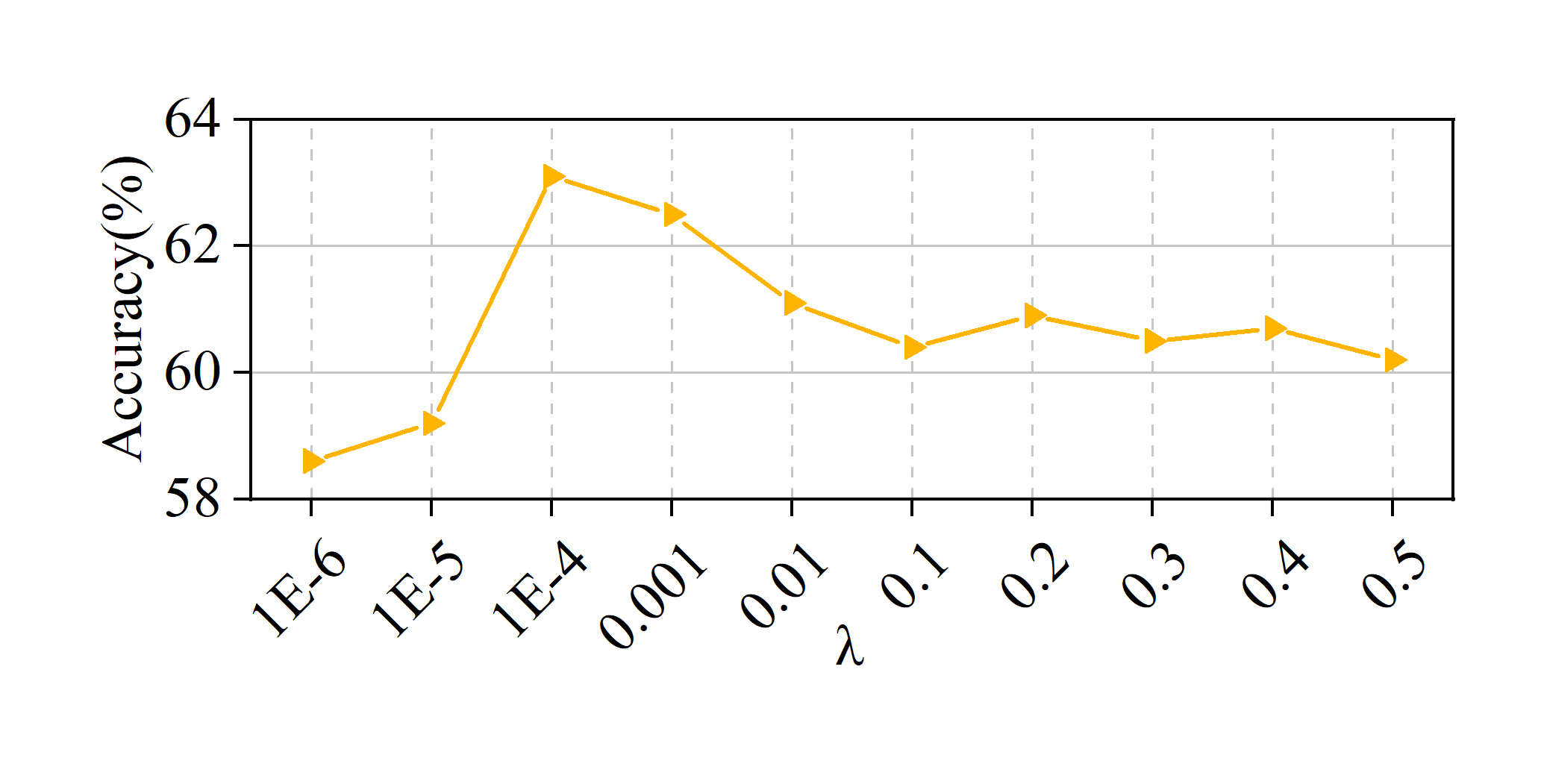}
            \caption{ABIDE}
        \end{subfigure}
        \begin{subfigure}[b]{0.48\textwidth} 
            \centering
            \includegraphics[width=\textwidth]{  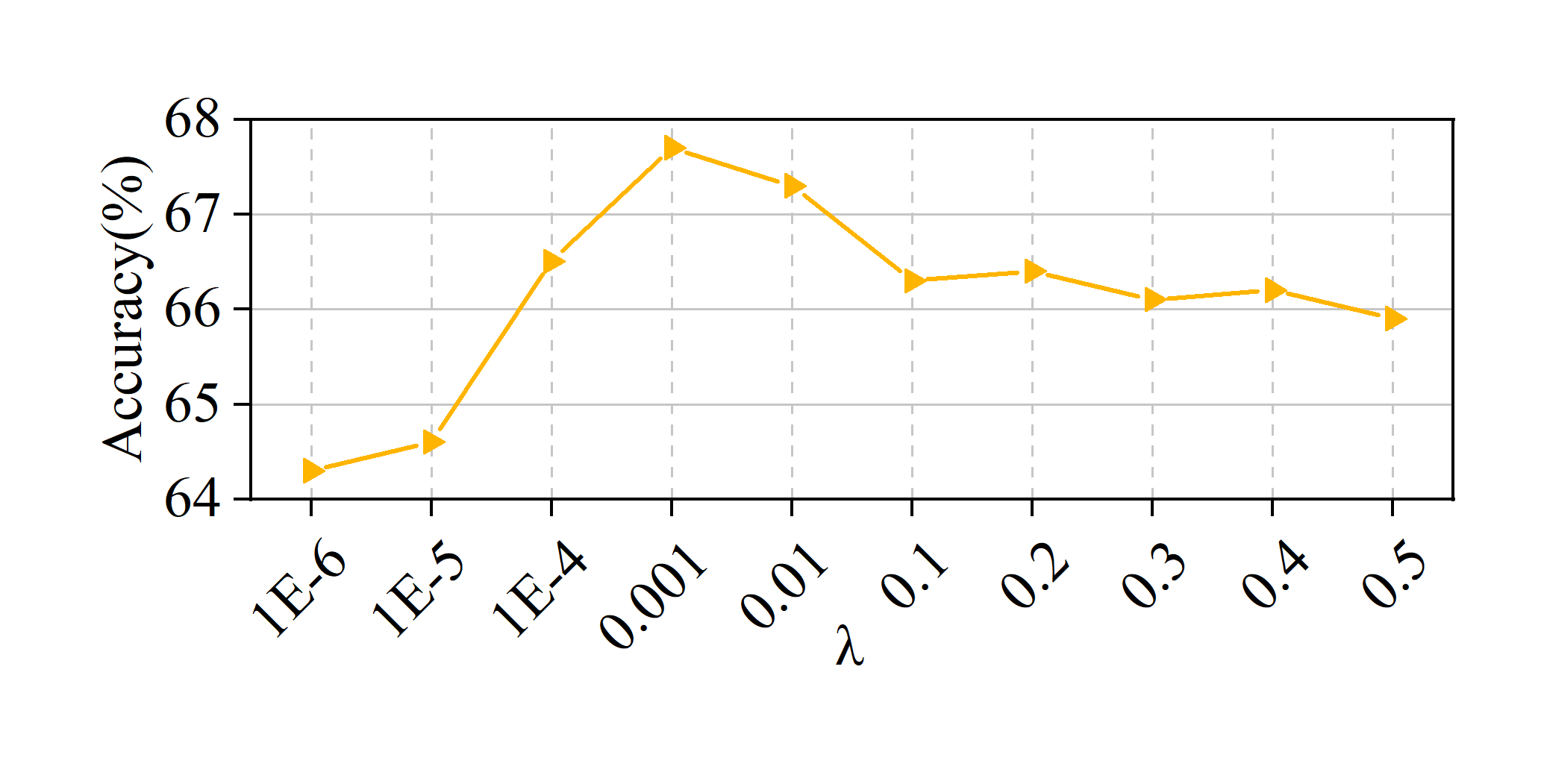}
            \caption{ADHD}
        \end{subfigure}
        \caption{Accuracy (\%) of SAM-BG with different values of $\lambda$ in Equation \ref{Loss_CCA}.}
        \label{fig: lambda}
    \end{figure}
    
    \paragraph{Hyperparameter Tuning Study.}

    We conduct a sensitivity analysis on two key hyperparameters: $\beta$ and $\lambda$. First, $\beta$ controls the sparsity regularization during the masker pre-training phase. We vary $\beta$ in the range \([1\text{e}^{-5}, 1.0]\) and evaluate its impact on classification performance for ABIDE and ADHD. As shown in Figure~\ref{fig: beta}, small values of $\beta$ result in weak regularization, leading to substructures with redundant or noisy edges. Conversely, large $\beta$ values over-prune the graph, potentially removing important connections. A balanced choice of $\beta=0.01$ for ABIDE and $\beta=0.001$ for ADHD achieves stable and high accuracy. 
    Second, we examine $\lambda$, which weights the decorrelation constraint during representation learning to promote semantic diversity between augmented views. As illustrated in Figure~\ref{fig: lambda}, a small $\lambda$ fails to prevent feature collapse, while a large $\lambda$ over-emphasizes diversity, impairing invariance learning. Optimal performance is achieved with $\lambda=1\text{e}^{-4}$ for ABIDE and $\lambda=0.001$ for ADHD, balancing diversity and consistency across views.

    \begin{figure}[t!]

        \centering
        
        \begin{subfigure}[b]{0.48\textwidth} 
            \centering
            \includegraphics[width=\textwidth]{  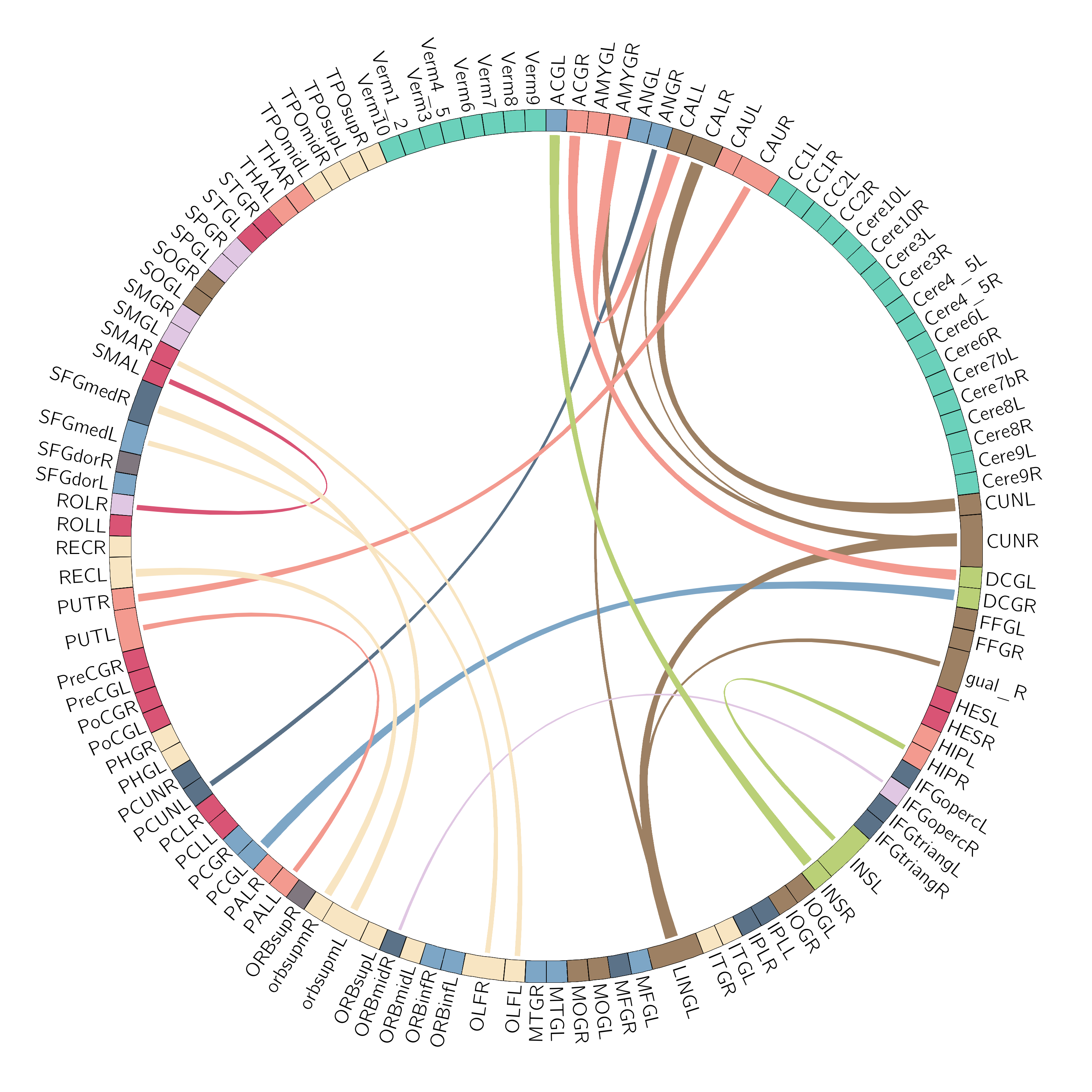}
            \caption{ASD}
        \end{subfigure}
        \begin{subfigure}[b]{0.48\textwidth} 
            \centering
            \includegraphics[width=\textwidth]{  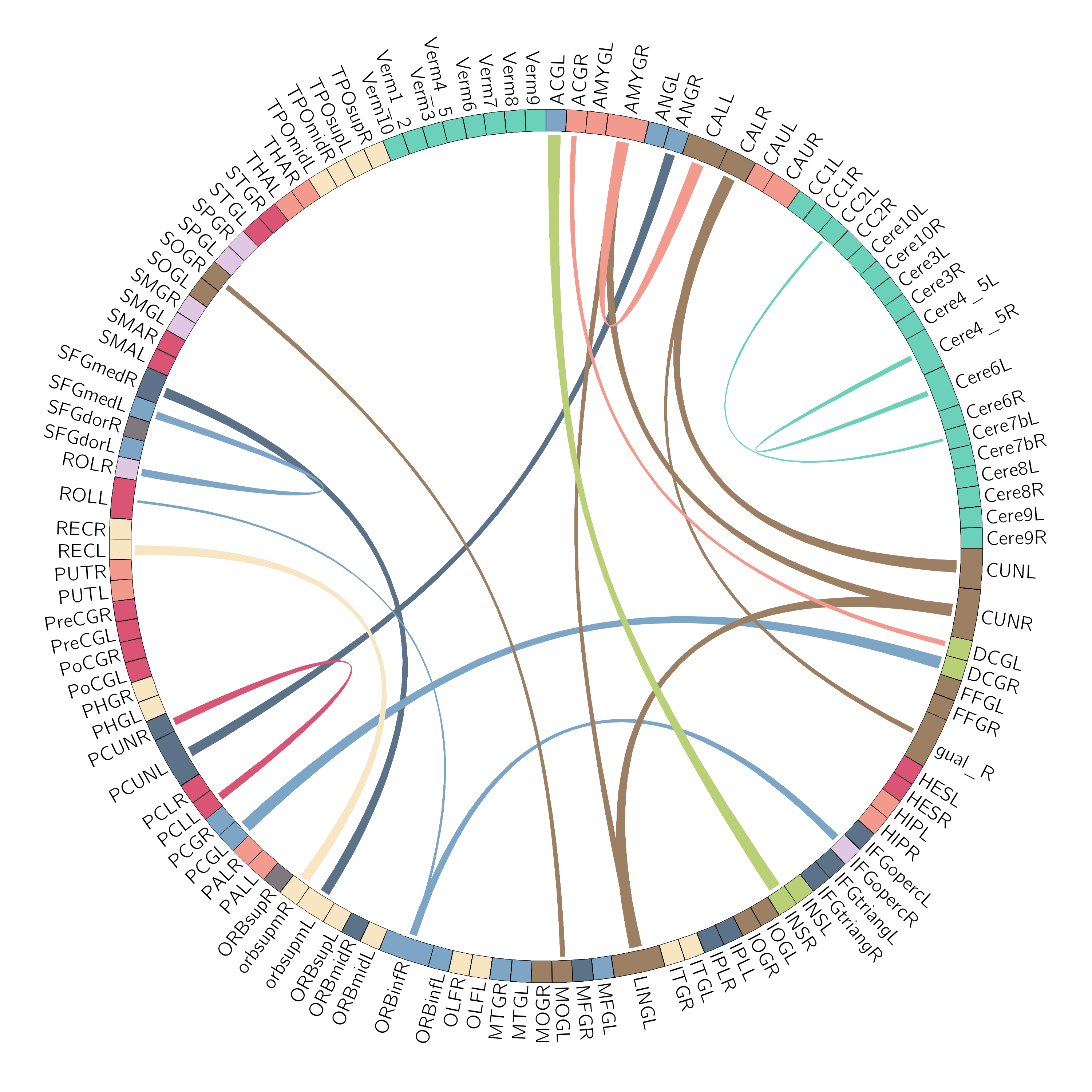}
            \caption{ADHD}
        \end{subfigure}

        \caption{Explainable graph connections in brain networks of patients with ASD and ADHD. The colors of brain networks are described as visual network (\textcolor{VN}{\textbf{VN}}), somatomotor network (\textcolor{SMN}{\textbf{SMN}}), dorsal attention network (\textcolor{DAN}{\textbf{DAN}}), ventral attention network (\textcolor{VAN}{\textbf{VAN}}), limbic network (\textcolor{LIN}{\textbf{LIN}}), frontoparietal network (\textcolor{FPN}{\textbf{FPN}}), default mode network (\textcolor{DMN}{\textbf{DMN}}), cerebellum (\textcolor{CBL}{\textbf{CBL}}) and subcortical network (\textcolor{SBN}{\textbf{SBN}}), respectively.} \label{fig: explain}
       \vspace{-10pt}
    \end{figure}

    \paragraph{Model Explainability Analysis.}

    The edge masker in SAM-BG improves representation learning and interpretability by identifying discriminative substructures that reflect abnormal brain connectivity patterns in psychiatric disorders. To demonstrate this, Figure~\ref{fig: explain} visualizes the top 20 connections from patients’ adjacency matrices, with nodes color-coded by nine functional networks and edge thickness indicating connection strength. In the ABIDE dataset, our method identified increased internal connectivity within the VN and enhanced connectivity between the LIN and other networks, including the DMN and FPN, reflecting heightened sensory sensitivity and over-processing of emotional inputs in autism \cite{eggebrecht2017joint, abbott2018repetitive}. In the ADHD dataset, the model highlights hyperconnectivity in the VAN, particularly in the right hemisphere, along with elevated internal connectivity within the CBL. These patterns are consistent with evidence linking the cerebellum to attention control and the VAN to stimulus-driven attention, supporting their relevance in ADHD~\cite{lin2021functional}. Interestingly, both datasets show increased VN connectivity, suggesting a shared neural signature that may underlie their frequent comorbidity. These results demonstrate that SAM-BG can uncover meaningful, explainable connectivity patterns, offering improved diagnostic insights into psychiatric disorders.

\section{Conclusion}

We propose SAM-BG, a novel framework that leverages structural priors to enable biologically meaningful brain graph representation learning under limited supervision. SAM-BG first trains an edge masker on a small labeled subset to extract salient connections, capturing key structural semantics. It then generates structure-aware augmentations by perturbing only non-essential edges, preserving semantic integrity for robust SSL. Experimental results demonstrate that SAM-BG substantially improves psychiatric disorder detection in low-data scenarios. Moreover, it provides explainable insights into brain connectivity patterns, underscoring its strong potential for real-world neuroscience applications.

\bibliographystyle{splncs04}
\bibliography{mybib}
\end{document}